\documentclass[11pt,a4paper]{article}
\usepackage[hyperref]{emnlp2020}
\usepackage{times}
\usepackage{latexsym}

\usepackage{enumitem}% http://ctan.org/pkg/enumitem
\usepackage{hyperref}
\usepackage{url}
\usepackage{tabularx, booktabs}
\usepackage{multirow}
\usepackage{xspace}
\usepackage{amsfonts}
\usepackage{amsmath}
\usepackage{amssymb}
\usepackage{graphicx}
\usepackage{subfigure}

\usepackage{microtype}

\aclfinalcopy % Uncomment this line for the final submission
\newcommand{\MBERT}{\mbox{\sf M-BERT}\xspace}
\newcommand{\EBERT}{\mbox{\sf E-MBERT}\xspace}

\newcommand{\BBERT}{\mbox{\sf B-BERT}\xspace}
\newcommand{\BERT}{\mbox{\sf BERT}\xspace}

\newcommand{\EXTEND}{\mbox{\sf Extend}\xspace}
\newcommand{\Extend}{\mbox{\sf Extend}\xspace}

\newcommand{\E}{\mbox{\sf E}\xspace}

\newcommand{\smallsection}[1]{{\noindent\textbf{#1.}}}

\newcommand*\samethanks[1][\value{footnote}]{\footnotemark[#1]}

\title{Extending Multilingual BERT to Low-Resource Languages}

\author{
Zihan Wang\thanks{Equal Contribution; most of this work was done while the author interned at the University of Pennsylvania.}\\
University of Illinois Urbana-Champaign \\
Urbana, IL 61801, USA \\
\texttt{zihanw2@illinois.edu} \\
\And
Karthikeyan K \samethanks\\
Indian Institute of Technology Kanpur \\
Kanpur, Uttar Pradesh 208016, India \\
\texttt{kkarthi@cse.iitk.ac.in} \\
\AND
Stephen Mayhew\thanks{This work was done while the author was a student  at the University of Pennsylvania.} \\
Duolingo \\
Pittsburgh, PA, 15206, USA \\
\texttt{stephen@duolingo.com} \\
\And
Dan Roth \\
University of Pennsylvania \\
Philadelphia, PA 19104, USA \\
\texttt{danroth@seas.upenn.edu} \\
}

\date{}

\begin{document}
\maketitle
\begin{abstract}
    % \drc{
Multilingual BERT (\MBERT) has been a huge success in both supervised and zero-shot cross-lingual transfer learning. However, this success has focused only on the top 104 languages in Wikipedia that it was trained on. In this paper, we propose a simple but effective approach to {\em extend} \MBERT (\EBERT) so that it can benefit any new language, and show that our approach benefits languages that are already in \MBERT as well. We perform an extensive set of experiments with Named Entity Recognition (NER) on 27 languages, only 16 of which are in \MBERT, and show an average increase of about 6\% F$_1$ on languages that are already in \MBERT and 23\% F$_1$ increase on new languages. 
% }

%Multilingual BERT (\MBERT) has been a huge success in both supervised and zero-shot cross-lingual transfer learning. However, this success of \MBERT is more focused to only the top 104 languages in Wikipedia that it is trained on. In this paper, we propose a simple but effective approach to include any new language to \MBERT. Our approach is also beneficial for languages that are already in \MBERT. We perform an extensive set of experiments with Named Entity Recognition (NER) task on 27 languages -- 16 of them are in \MBERT, and the other 11 not --  and the results show that on an average we increase about 6\% F$_1$ on languages that are already in \MBERT and 23\% F$_1$ on new languages. 

\end{abstract}
\section{Introduction}

% \dr{You will need to refer to the corpus as the LORELEI corpus and cite \cite{strassel-tracey-2016-lorelei}}

Recent works~\cite{wu-dredze-2019-beto, karthikeyan2020cross} have shown the zero-shot cross-lingual ability of \MBERT~\cite{devlin2018multi} on various semantic and syntactic tasks -- just fine-tuning on English data allows the model to perform well on other languages.  Cross-lingual learning is imperative for low-resource languages (LRL), such as Somali and Uyghur, as obtaining supervised training data in these languages is particularly hard. However, \MBERT is not pre-trained with these languages, thus limiting its performance on them. Languages like Oromo, Hausa, Amharic and Akan are spoken by more than 20 million people, yet \MBERT does not cover these languages.  Indeed, there are about 4000\footnote{\url{https://www.ethnologue.com/enterprise-faq/how-many-languages-world-are-unwritten-0}} languages written by humans, of which \MBERT covers only the top 104 languages (less than 3\%). 

One of the approaches to use the idea of \MBERT for languages that are not already present is to train a new \MBERT from scratch. However, this is extremely time-consuming and expensive: training BERT-base itself takes about four days with four cloud TPUs~\cite{devlin-etal-2019-bert}, so training \MBERT should take even more time\footnote{The exact training time was not reported.}. Alternatively, we can train Bilingual BERT (\BBERT)~\cite{karthikeyan2020cross}, which is more efficient than training an \MBERT. However, one major disadvantage of \BBERT is that we can not use supervised data from multiple languages, even if it is available.

\begin{figure}[t]
    \centering
    \includegraphics[width=1\linewidth]{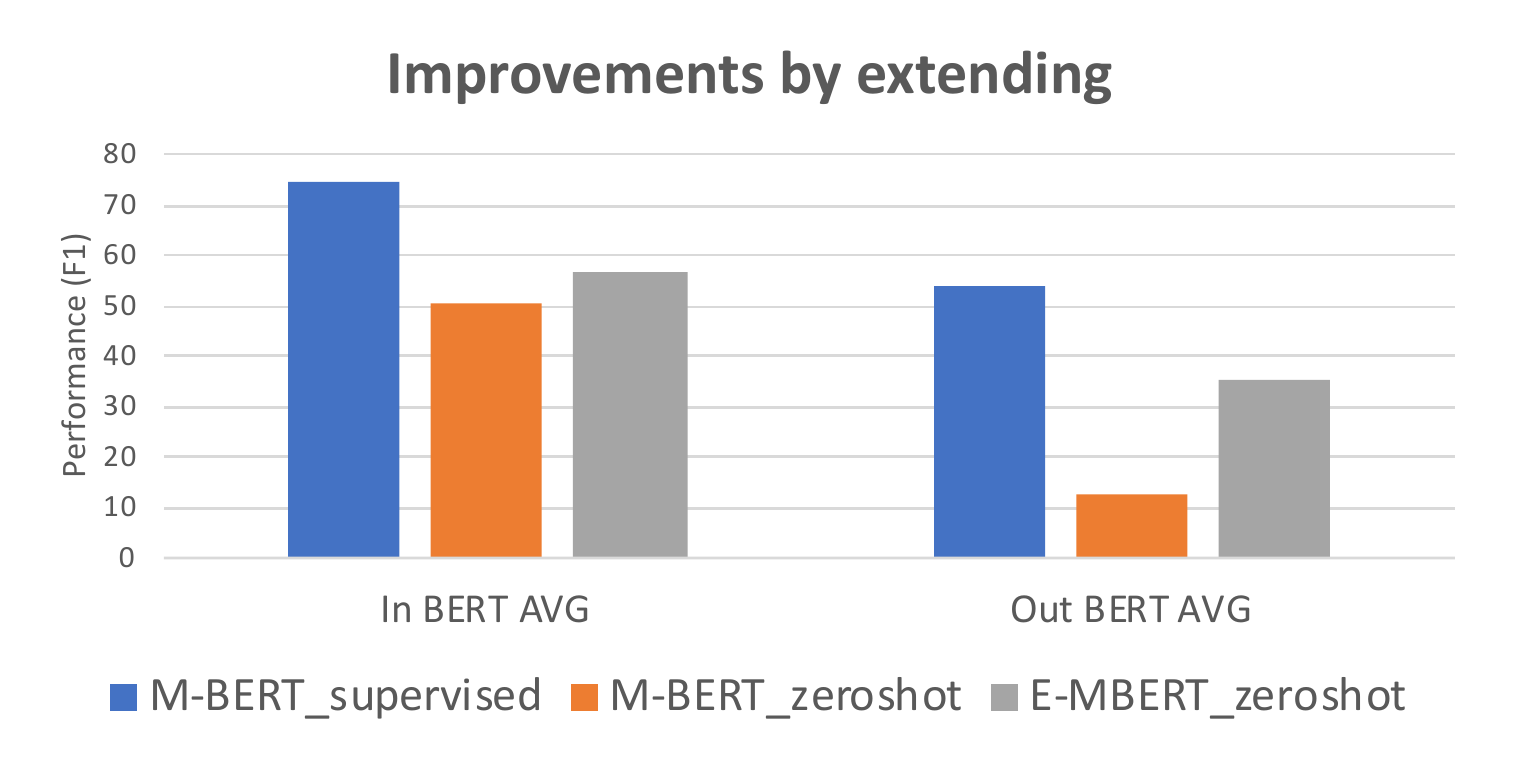}
    % \caption{Average scores across language in \MBERT or out \MBERT on NER with or without our extending approach and with or without supervised training data.}
    \caption{\textbf{Comparison between \MBERT and our proposed approach \EBERT: } We report averaged zero-shot NER performance on 16 languages that are already {\em in} \MBERT %IN BERT 
    and 11 new language that are {\em out} of \MBERT.
    %OUT of BERT. 
    We also report average of \MBERT's performance with supervised NER data as a upper-bound.}
    %\dr{Font in the figure is too small, especially the legend; also, change to M-BERT; extend-->Extended}}
    \label{fig:avg_mbert}
    \vspace{-0.4cm}
\end{figure}

To accommodate a language that is not in \MBERT, we propose an efficient approach \EXTEND that adapts \MBERT to the language. \EXTEND works by enlarging the vocabulary of \MBERT to accommodate the new language and then continue pre-training on this language. Our approach consumes less than $7$ hours to train with a single cloud TPU. 
%It is efficient mainly because instead of randomly initializing, we initialize most of the weights (not all weights) from the pre-trained \MBERT.
% Through experiments we also show that \EXTEND is beneficial for languages that are already in \MBERT.
% In this work, we propose a very efficient approach \EXTEND, where we can add a new language to \MBERT; that is, we can add a language like Somali to \MBERT, so that the new \MBERT can work well on Somali. \EXTEND works by extending the vocabulary to accommodate the new language and then continue training on \MBERT on the language.

We performed comprehensive experiments on NER task with 27 languages of which 11 languages are not present in \MBERT. From Figure~\ref{fig:avg_mbert} we can see that our approach performs  significantly better than \MBERT when the target language is out of the 104 languages in \MBERT. Even for high-resource languages that are already in \MBERT, our approach is still superior.

The key contributions of the work are (i) We propose a simple yet novel approach to add a new language to \MBERT (ii) We show that our approach improves over \MBERT for both languages that are in \MBERT and out of \MBERT (iii) We show that, in most cases, our approach is superior to training \BBERT from scratch. Our results are reproducible and we will release both the models and code.

% To address the issue of lack of training data in Low Resource Languages, researchers propose many cross-lingual approaches. Amongst these methods, \MBERT has gained a great deal of attention because of it requiring only text corpus, and no cross-lingual input. While \MBERT is trained on the top 104 languages in Wikipedia (ranked by number of articles), there are about 4000\footnote{\url{https://www.ethnologue.com/enterprise-faq/how-many-languages-world-are-unwritten-0}} languages that are written by humans, indicating that \MBERT covers less than $3\%$ of them.

% In this paper, we focus on a specific task of cross-linguality, namely Cross-Lingual Named Entity Recognition. To address a language that is not contained inside \MBERT, we propose \EXTEND, which is to extend the current \MBERT to adapt the language by continuing the \BERT pre-training process on the language. 

% With extensive experiments, we show that our method brings significant improvements across almost all languages we experiment with comparing to \MBERT, even with languages that are already in \MBERT.

% Figure~\ref{fig:avg_mbert} shows a summary of how our extending approach compares with \MBERT. It can be seen that our approach results in significant performance improvements over \MBERT when the target language is out of the 104 languages in \MBERT. However, even for high resource languages that are inside \MBERT, our approach is still superior.
\section{Related works}

Cross-lingual learning has been a rising interest in NLP. For example, BiCCA~\cite{faruqui-dyer-2014-improving}, LASER~\cite{DBLP:journals/tacl/ArtetxeS19} and XLM~\cite{DBLP:conf/nips/ConneauL19}. Although these models have been successful, they need some form of cross-lingual supervision such as a bilingual dictionary or parallel corpus, which is particularly challenging to obtain for low-resource languages. Our work differ from above as we do not require such supervision. While other approaches like MUSE~\cite{lample2018word} and VecMap~\cite{artetxe-etal-2018-robust} can work without any cross-lingual supervision, \MBERT already often outperforms these approaches~\cite{karthikeyan2020cross}.

~\citet{schuster-etal-2019-cross-lingual} has a setting of continuing training similar to ours. However, their approach focus more on comparing between whether \BBERT (JointPair) learns cross-lingual features from overlapping word-pieces, while ours focus more on improving \MBERT on target languages, and addresses the problem of missing word-pieces. We show that our \Extend method works well on \MBERT, and is better than \BBERT in several languages, whereas their method (MonoTrans) has a similar performance as \BBERT. This together implies that our \Extend method benefits from the multilinguality of the base model (\MBERT vs \BERT).

% , and one of the most popular approaches before the advent of \MBERT is to train bilingual or multilingual word embeddings~\cite{upadhyay-etal-2016-cross,AMTLDS16}

% , Bilingual Skip-Gram Model (BiSkip)~\cite{luong-etal-2015-bilingual}, and Bilingual Compositional Model (BiCVM)~\cite{hermann-blunsom-2014-multilingual}.

% Further, recent advances in the direction of BERT like models such as RoBERTa~\cite{liu2019roberta}, XLNet~\cite{yang2019xlnet} has been very promising and these approaches can be adapted to multilingual case as well.  

\section{Background}
\subsection{Multilingual BERT (M-BERT) }

\MBERT is a bi-directional transformer language model pre-trained with Wikipedia text of top 104 languages -- languages with most articles in Wikipedia. \MBERT uses the same pre-training objective as BERT -- masked language model and next sentence prediction objectives~\cite{devlin-etal-2019-bert}. Despite not being trained with any specific cross-lingual objective or aligned data, \MBERT is surprisingly cross-lingual. For cross-lingual transfer, \MBERT is fine-tuned on supervised data in high-resource languages like English and tested on the target language.

\subsection{Bilingual BERT (B-BERT) }
\BBERT is trained in the same way as \MBERT except that it contains only two languages -- English and the target language. Recent works have shown the cross-lingual effectiveness of \MBERT~\cite{pires-etal-2019-multilingual, wu-dredze-2019-beto}, and \BBERT~\cite{karthikeyan2020cross} on NER and other tasks.
\section{Our Method: Extend}

In this section, we discuss our training protocol \EXTEND which works by extending the vocabulary, encoders and decoders to accommodate the target language and then continue pre-training on this language. 

% In this section, we present a novel training protocol called \EXTEND, where we can add any new language to \MBERT. 

% Our approach is beneficial for not only the new languages but also languages that already exist in M-BERT. Experiments show that our method performs significantly better than M-BERT, in both zero-shot and supervised scenarios.  

Let the size of \MBERT's vocabulary be $V_{mbert}$ and the embedding dimension be $d$. We first create the vocabulary with the monolingual data in the target language following the same procedure as \BERT, and filter out all words that appear in \MBERT's vocabulary. Let the size of this new vocabulary be $V_{new}$. Throughout the paper, we set $V_{new} = 30000$. Then, we append this new vocabulary to \MBERT's vocabulary.   
% We start with an initial \MBERT, which is already pre-trained with several languages, and then create the vocabulary for the new language following the same procedure as the \MBERT (It involves only one language, so it is same as \BERT's vocabulary creation). 
%We append this new-vocabulary (new-vocabulary refers to only the entries that are not already present in \MBERT's vocabulary) to the initial \MBERT's vocabulary to create our final vocabulary. Let the size of \MBERT's vocabulary and new-vocabulary be $V_{mbert}$ and $V_{new}$, respectively.  
We \textit{extend} the encoder and decoder weights of the \MBERT model so that it can encode and decode the new-vocabulary. That is, we extend the \MBERT's encoder matrix of size $V_{mbert} \times d$ with a matrix of size $V_{new} \times d$ , which is initialized following \MBERT's procedure, to create an extended encoder of size $\left(V_{mbert} + V_{new}\right) \times d$; we do similar extension for decoder. Note that M-BERT uses weight-tying, hence the decoder is the same as the encoder, except it has an additional bias. 

We then continue pre-training with the monolingual data of the target language. Note that except for the newly appended part of encoder and decoder, we initialize all weights with \MBERT's pre-trained weight. We call the trained model model \EBERT.

\section{Experiments}
\begin{figure*}[t]
    \centering
    \includegraphics[width=1\linewidth]{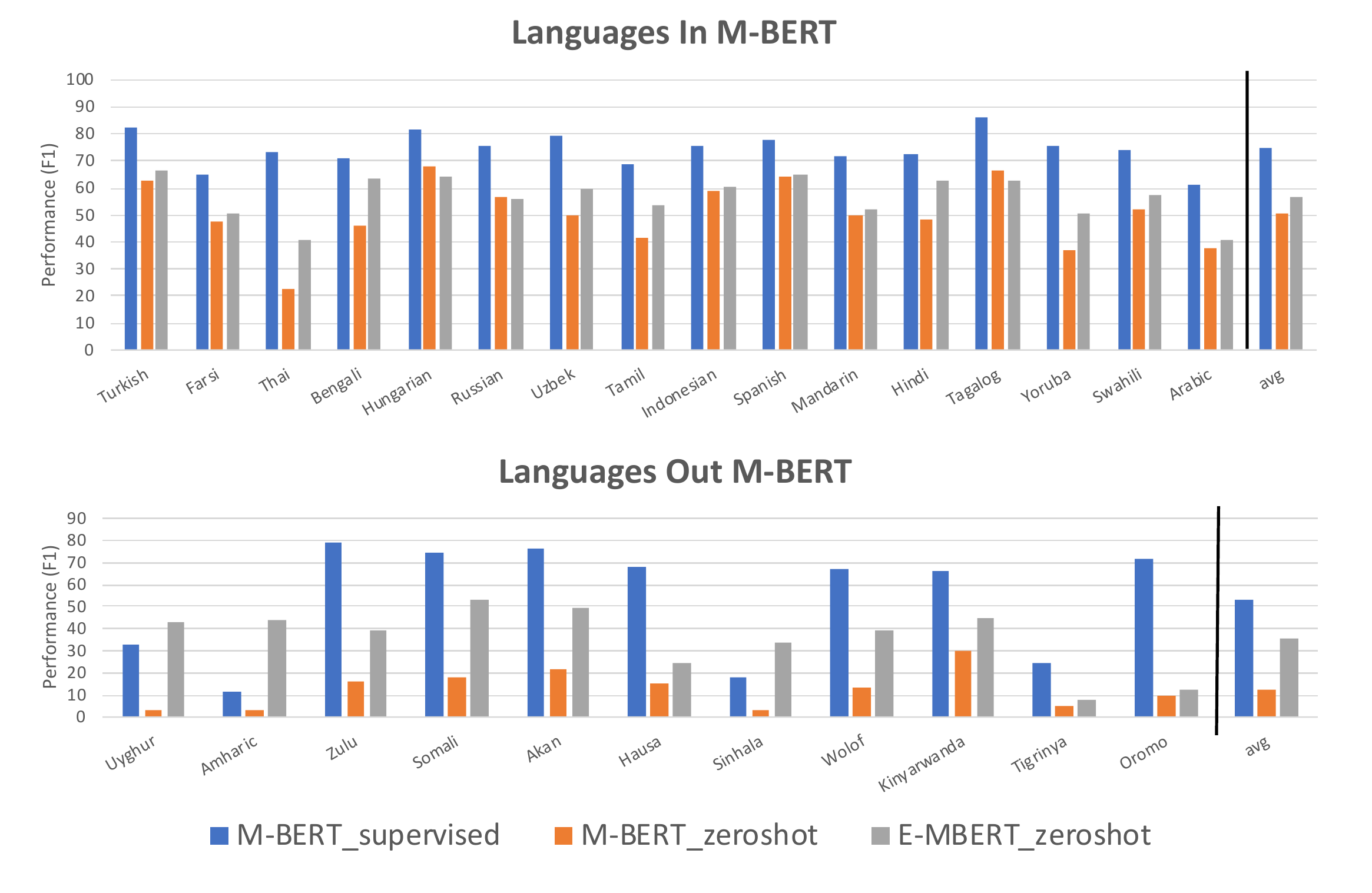}
    % \caption{Scores across language in \MBERT or out of \MBERT on NER with or without our extending approach and with or without supervised training data.}
    \caption{\textbf{Comparison between M-BERT and E-MBERT: } We compare zero-shot cross-lingual NER performance (F1 score) on \MBERT and \EXTEND using 27 languages. The languages are ordered by amount of text data in LORELEI.  We also report the \MBERT's supervised performance as a benchmark to compare.}
    \label{fig:all_mbert}
    \vspace{-0.4cm}
\end{figure*}
\subsection{Experimental Settings}
\smallsection{Dataset} Our text corpus and NER dataset are from LORELEI~\cite{strassel2016lorelei}. We use the tokenization method from \BERT to preprocess text corpuses. For zero-shot cross-lingual NER, we evaluate the performance on the whole annotated set; for supervised learning, since we just want an understanding of a upper-bound, we apply cross validation to estimate the performance: each fold is evaluated by a model trained on the other folds, and the average F$_1$ is reported.

\smallsection{NER Model} We use a standard Bi-LSTM-CRF~\cite{ma-hovy-2016-end,lample-etal-2016-neural} framework and use AllenNLP~\cite{gardner-etal-2018-allennlp} as our toolkit. The scores reported in NER is the F$_1$ score averaged across five runs with different random seeds. 

% Since we want to compare the performance of different \BERT training procedures, we use the same standard Bi-LSTM-CRF framework on top of static contextualized embeddings of a \BERT model of interest. The scores reported in NER is the $F_1$ score averaged across five runs. 

\smallsection{BERT training} While extending, we use a batch size of 32 and a learning rate of 2e-5, which \BERT suggests for fine-tuning, and we train for 500k iterations. Whereas for \BBERT we use a batch size of 32 and learning rate of 1e-4 and train for 2M iterations. We follow \BERT setting for all other hyperparameters.

\subsection{Comparing between E-MBERT and M-BERT}

\label{sec:exp_cmp_mbert}

We compare the cross-lingual zero-shot NER performance of \MBERT and \EBERT. We train only with supervised LORELEI English NER data. We also report the performance of \MBERT with supervision on the target language, which allows us to get a reasonable ``upper-bound'' on the dataset. From Figure~\ref{fig:all_mbert}, we can see that in almost all languages, \EBERT outperforms \MBERT irrespective of whether they exist or do not exist in \MBERT.

It is clear that \EBERT performs better than \MBERT when the language is not present; however, it is intriguing that \EBERT improves over \MBERT when the language is already present in \MBERT. We attribute this improvement in performance to three reasons 
\begin{itemize}[leftmargin=*,nosep]
    \item Increased vocabulary size of target language --  Since most languages have a significantly smaller Wikipedia data than English, they have a fewer vocabulary in \MBERT, our approach eliminates this issue. Note that it may not be a good idea to train single \MBERT with larger vocabulary sizes for every language, as this will create a vast vocabulary (a few million).
    \item \EBERT is more focused on the target language, as during the last 500k steps, it is optimized to perform well on it.
    \item Extra monolingual data -- More monolingual data in the target language can be beneficial. 
\end{itemize}

% (i) Increased vocabulary size of target language --  Since most languages have a significantly smaller Wikipedia data than English, they have a fewer vocabulary in \MBERT, our approach eliminates this issue. Note that it may not be a good idea to train single \MBERT with larger vocabulary sizes for every language, as this will create a vast vocabulary (a few million). (ii) \EBERT is more focused on the target language, as during the last 500k steps, it is optimized to perform well on it. (iii) Extra monolingual data -- More monolingual data in the target language can be beneficial. 
%However, this is not the only reason; our training protocol itself seems to be useful, refer table~\ref{tbl:wiki}. 

%The zeroshot setting represents a practical situation where we don't have annotated training data in a target language, so we transfer knowledge from a high resource language.
\begin{table}[!htbp]
\centering
\begin{tabular}{l c | c c} 
\toprule
\textbf{Lang} & \MBERT &  \E w/ Lrl & \E w/ Wiki \\
\midrule
Russian & 56.56 & 55.70 & 56.64\\
Thai & 22.46 & 40.99 & 38.35 \\
Hindi & 48.31 & 62.72 & 62.77 \\
\bottomrule
\end{tabular}
\caption{Performance of \MBERT, \EXTEND with LORELEI data and \EXTEND with Wikipedia data.}
\label{tbl:wiki}
\vspace{-0.4cm}
\end{table}

\subsection{Extend without extra data}
The effectiveness of \EBERT may be partially explained by the extra monolingual data the model is trained on. To explore the performance of \EBERT without this extra training data, we \EXTEND with using Wikipedia data, which is used in \MBERT. From Table~\ref{tbl:wiki}, we can see that even without additional data, \EBERT's performance does not degrade.

\begin{table}[!htbp]
\centering
\begin{tabular}{l c c} 
\toprule
\textbf{Lang} & \BBERT &  \EXTEND \\
\midrule
Somali & 51.18 & \textbf{53.63}\\
Amharic & 38.66 & \textbf{43.70}\\
Uyghur & 21.94 & \textbf{42.98}\\
Akan & 48.00 & \textbf{49.02} \\
Hausa & \textbf{26.45} & 24.37 \\
Wolof & \textbf{39.92} & 39.70 \\
Zulu & \textbf{44.08} & 39.65 \\
Tigrinya & 6.34 & \textbf{7.61} \\
Oromo & 8.45 & \textbf{12.28} \\
Kinyarwanda & \textbf{46.72} & 44.40 \\
Sinhala & 16.93 & \textbf{33.97} \\
\bottomrule
\end{tabular}
\caption{\textbf{Comparison between B-BERT and E-MBERT: } We compare \BBERT vs \EBERT training protocols. Both the models uses same target language monolingual data. We compare the performances on languages that are not in \MBERT, so that \EBERT doesn't make use of \MBERT's additional Wikipedia data.}
\label{tbl:bbert}
\vspace{-0.4cm}
\end{table}

\subsection{Comparing between E-MBERT and B-BERT}

Another way of addressing \MBERT on unseen languages is to completely train a new \MBERT. Restricted by computing resources, it is often only feasible to train on both the source and the target, hence a bilingual BERT (\BBERT). Both \EBERT with \BBERT uses the same text corpus in the target language; for \BBERT, we subsample English Wikipedia data. We focus only on languages that are not in \MBERT so that \EBERT will not have an advantage on the target language because of data from Wikipedia. Although the English corpus of \BBERT is different from \EBERT, the difference is marginal considering its size. Indeed we show that \BBERT and \EBERT have similar performance on English NER, refer Appendix~\ref{sec:appendix_english} and Appenddix~\ref{sec:appendix_bbert}.

From Table~\ref{tbl:bbert}, we can see that \EBERT often outperforms \BBERT. Moreover, \BBERT is trained for 2M steps for convergence, while \EBERT requires only 500k steps. We believe that this advantage comes for the following reason: \EBERT makes use of the better multilingual model \MBERT, which potentially contains languages that help transfer knowledge from English to target, while \BBERT can only leverage English data. For example, in the case of Sinhala and Uyghur, a comparatively high-resource related language like Tamil and Turkish in \MBERT can help the \EBERT learn Sinhala and Uyghur better.

\subsection{Rate of Convergence}
\begin{figure}[t]
    \centering
    \includegraphics[width=1\linewidth]{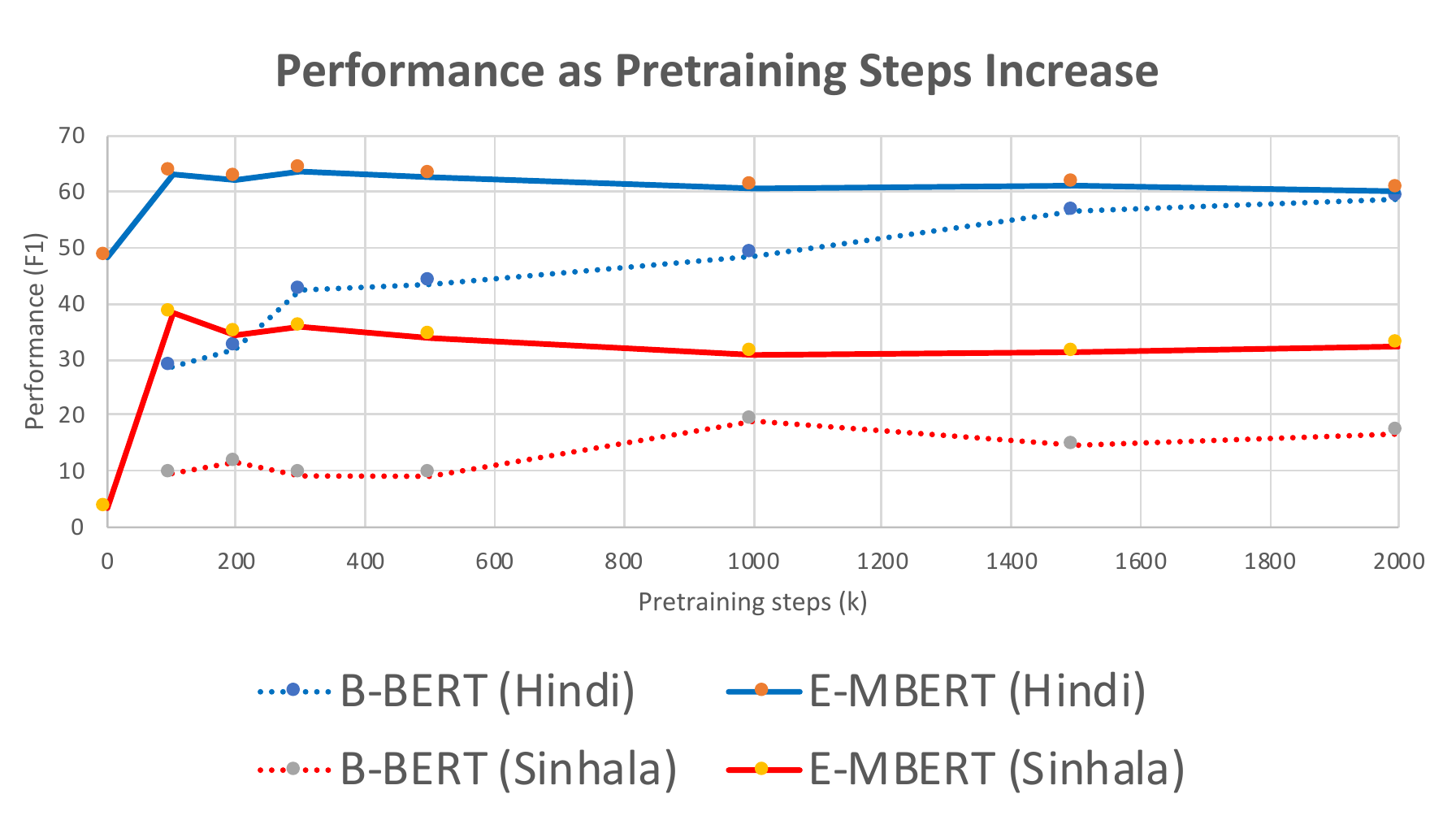}
    \caption{Performance of \BBERT and \EXTEND as number of pre-training steps increase}
    \label{fig:convergence}
    \vspace{-0.6cm}
\end{figure}

In this subsection, we study the convergence rate of \EBERT and \BBERT. We evaluate these two models on two languages, Hindi (in \MBERT) and Sinhala (not in \MBERT), and report the results in Figure~\ref{fig:convergence}. We can see that \EBERT is able to converge within just 100k steps, while for \BBERT, it takes more than 1M steps to converge. This shows that \EBERT is much more efficient than \BBERT.

\subsection{Performance on non-target languages}
Our \EXTEND method results in the base model (\MBERT) to focus on the target language, and naturally this degrades performance on the other languages that are not the target language. We report the performance of Hindi and Sinhala \EBERT models evaluated on the other languages in Appendix~\ref{sec:appendix_data}.

% In the case of Sinhala and Uyghur, \EXTEND performs significantly better than \BBERT. One of the potential reasons this improvement could be the existence of comparatively high-resource related language in \MBERT -- Tamil (for Sinhala) and Uzbek and Turkish (for Uyghur).

% As a parallel way of analyzing \EBERT without extra data is to \EXTEND with data from the top 104 Wikipedia. In Table~\ref{tbl:wiki}, we report the performance on Russian, Thai and Hindi. We show that even if we do not introduce new data, \EXTEND still leads to a similar performance as if task domain related corpus is used. 

% To explore the performance of \EBERT without extra training data, we retrain a \BBERT with only data on English and target language. We focus on languages that do not exist in \MBERT, therefore \EXTEND will not have an advantage on the target language because of data from Wikipedia.Note that the English corpus is different, however,  In Table~\ref{tbl:bbert}, we report the results on the target languages. 

% \subsection{\EXTEND \BBERT}
% \input{tbl_extend_bbert.tex}
% The idea of \EXTEND can be applied to models beyond \MBERT. In Table~\ref{tbl:extend_bbert} we show that when \EXTEND is applied to \BBERT, we also witness performance improvements, despite the exact same training data is used. This shows the generality of our training protocol.

\section{Conclusions and Future work}
In this work, we propose \EXTEND that deals with languages not in \MBERT. Our method has shown great performance across several languages comparing to \MBERT and \BBERT.

While \EXTEND deals with one language each time, it would be an interesting future work to extend on multiple languages at the same time. Furthermore, instead of randomly initialising the embeddings of new vocabulary, we could possibly use alignment models like MUSE or VecMap with bilingual dictionaries to initialize. We could also try to apply our approach to better models like RoBERTa~\cite{liu2019roberta} in multilingual case.

% Further, recent advances in the direction of BERT like models such as RoBERTa, XLNet~\cite{yang2019xlnet} has been very promising and these approaches can be adapted to multilingual case as well.  

% There are quite a few Our work is focused on extending to a single language, while it is certainly possible to extend to multiple.

% Our current approach extends only one language at a time, it would be an interesting future work to analyze extending with multiple languages together. 

% Another immediate extension would be to train a Multilingual RoBERTa and apply \EXTEND protocol. 

% Although \MBERT is imperative for low-resource language when there is no cross-lingual supervision  (like bilingual dictionary) is available, it would be wise to use dictionaries or parallel data~\cite{conneau2019cross} when available. Particularly, \EXTEND is suitable to use with dictionaries -- instead of randomly initializing, we can carefully initialize the newly appended part of the encoder using a dictionary, may be using cross-lingual alignment methods like MUSE or VecMap. 

% Finally, it would interesting to analyze the \EXTEND on BERT-base, BERT-Large or ROBERTa~\cite{liu2019roberta}, XLNet~\cite{yang2019xlnet}, that is instead of starting with a multilingual model can we start with monolingual only models?

\bibliography{new,compact_anthology,cited}
\bibliographystyle{acl_natbib}
\appendix
\section{Appendices} \label{sec:appendix_english}

\subsection{Performance of \EBERT on English: }
The knowledge of \EBERT on English (source language) is not affected. From Table~\ref{tbl:eng-comp}, we can see that, except for few languages, the English performance of \EBERT is almost as good as \MBERT's. 

\subsection{Detailed data on all languages} \label{sec:appendix_data}

In Table~\ref{tbl:performance-comp}, we report the full result on
comparing \MBERT and \EBERT.
%the cross-lingual and supervised NER performance (F1 score) on all the 27 languages (Figure~\ref{fig:all_mbert} is drawn using this data). We can see that \EXTEND is significantly better than \MBERT  when the language is not present in \MBERT. Moreover, even when the language is already present in\MBERT, \EXTEND seems to improve over \MBERT; it improves on almost all the languages except Hungarian, Russian, and Tagalog.

We can also see that \EXTEND is not only useful for cross-lingual performance but also for useful for supervised performance (in almost all cases). 

We also notice that extending on one language hurts the transferability to other languages.

\subsection{Comparison between \BBERT and \EBERT: } \label{sec:appendix_bbert}
In Table~\ref{tbl:bibert-embert-eng-tar} we reported the performance of \EXTEND and \BBERT on both English as well as target. We can see that English performance of \BBERT is mostly better than \EXTEND. However, in most cases \EXTEND performs better on target language. This indicates that \EBERT does not have an unfair advantage on English.

\begin{table}[!htbp]
\centering
\begin{tabular}{l l } 
\toprule
\textbf{EXTEND Language} & \textbf{ E M-BERT}  \\
\midrule
\multicolumn{2}{c}{OUT OF BERT} \\
\midrule
Akan &  79.19 \\
Amharic & 78.36 \\
Hausa &  74.24 \\
Somali & 78.6 \\
Wolof &  78.11\\
Zulu &  79.32 \\
Uyghur & 77.76 \\
Tigrinya &  76.21 \\
Oromo & 76.06 \\
Kinyarwanda &  73.05 \\
Sinhala & 73.7 \\
\midrule
\multicolumn{2}{c}{IN BERT} \\
\midrule
Arabic & 77.67 \\
Bengali & 76.2 \\
Mandarin &  78.58 \\
Farsi &  77.57 \\
Hindi & 78.86 \\
Hungarian &  78.92\\
Indonesian &  80.93 \\
Russian & 80.87\\
Spanish &  81.15 \\
Swahili & 77.72\\
Tamil & 77.6 \\
Tagalog & 79.56\\
Thai & 78.21 \\
Turkish & 79.49 \\
Uzbek & 77.19 \\
Yoruba & 77.55 \\
\midrule
\textbf{M-BERT} & 79.37 \\
\bottomrule
\end{tabular}
\caption{\textbf{Performance on English: } We report the English NER performance of \MBERT as well as performance \EBERT.}
\label{tbl:eng-comp}
\end{table}

% \begin{table*}[!htbp]
% \centering
% \begin{tabular}{l l l l} 
% \toprule
% \textbf{Model} & INBERT &\textbf{ \E \MBERT}  & \textbf{ \BBERT} \\
% \midrule
% \textbf{M-BERT} & -  & 79.37 & -  \\
% \midrule
% \textbf{aka} (Akan) & False & 79.19 & 77.49 \\
% amh (Amharic) & False & 78.36 & 78.44\\
% \textbf{ara} (Arabic) & True & 77.67 & /\\
% \textbf{ben} (Bengali) & True & 76.2 & /\\
% \textbf{cmn} (Mandarin) & True & 78.58 & /\\
% \textbf{fas} (Farsi) & True & 77.57 & /\\
% hau (Hausa) & False & 74.24 & 80.13\\
% \textbf{hin} (Hindi) & True & 78.86 & /\\
% \textbf{hun} (Hungarian) & True & 78.92 & /\\
% \textbf{ind} (Indonesian) & True & 80.93 & /\\
% \textbf{rus} (Russian) & True & 80.87 & /\\
% som (Somali) & False & 78.6 & 79.17\\
% \textbf{spa} (Spanish) & True & 81.15 & /\\
% \textbf{swa} (Swahili) & True & 77.72 & /\\
% \textbf{tam} (Tamil) & True & 77.6 & /\\
% \textbf{tgl} (Tagalog) & True & 79.56 & /\\
% \textbf{tha} (Thai) & True & 78.21 & /\\
% \textbf{tur} (Turkish) & True & 79.49 & /\\
% \textbf{uzb} (Uzbek) & True & 77.19 & /\\
% wol (Wolof) & False & 78.11 & 81.01\\
% \textbf{yor} (Yoruba) & True & 77.55 & /\\
% zul (Zulu) & False & 79.32 & 81.82\\
% uig (uyghur) & False & 77.76 & 79.65\\
% tir (Tigrinya) & False & 76.21 & 80.35\\
% orm (Oromo) & False & 76.06 & 78.13\\
% kin (Kinyarwanda) & False & 73.05 & 79.37\\
% sin (Sinhala) & False & 73.7 & 80.04\\
% \bottomrule
% \end{tabular}
% \caption{English results}
% \label{tbl:eng-comp}
% \end{table*}

\begin{table*}[!htbp]
\centering
\begin{tabular}{l c c c c | c c | c c} 
\toprule
\multicolumn{9}{c}{In BERT} \\
\midrule
\textbf{Model} & \textbf{M-sup} & \textbf{M-zero}  & \textbf{E-sup} &  \textbf{E-zero} & Hindi & Sinhala & Corpus (M) & NER (k)\\
\midrule
\textbf{Arabic} & 61.14 & 37.56 & 61.97 & 40.83 & 19.2 & 16.72 & 0.19 &5.50 \\
\textbf{Bengali} & 71.29 & 46.18 & 84.44 & 63.49 &17.94 & 14.01 &10.19 &11.65 \\
\textbf{Mandarin} & 71.76 & 50.0 & 73.86 & 52.30 &8.88 & 24.64 &1.66 &8.05 \\
\textbf{Farsi} & 65.09 & 47.71 & 68.27 & 50.26 &22.38 & 20.44 &10.32 &4.38 \\
\textbf{Hindi} & 72.88 & 48.31 & 81.15 & 62.72 &62.72 & 18.0 &1.66 &6.22 \\
Hungarian & 81.98 & 68.26 & 82.08 & 64.36 &24.38 & 35.74 &10.09 &5.81 \\
\textbf{Indonesian} & 75.67 & 58.91 & 80.09 & 60.73 &29.5 & 37.89 &1.75 &6.96 \\
Russian & 75.60 & 56.56 & 76.51 & 55.70 &26.08 & 36.15 &10.07 &7.26 \\
\textbf{Spanish} & 78.12 & 64.53 & 78.14 & 64.75 &37.06 & 47.32 &1.68 &3.48 \\
\textbf{Swahili} & 74.26 & 52.39 & 81.9 & 57.21 &25.46 & 31.91 &0.29 &5.61 \\
\textbf{Tamil} & 68.55 & 41.68 & 77.91 & 53.42 &14.75 & 12.96 &4.47 &15.51 \\
Tagalog & 85.98 & 66.50 & 88.63 & 62.61 &34.73 & 42.16 &0.33 &6.98 \\
\textbf{Thai} & 73.58 & 22.46 & 86.40 & 40.99 &4.03 & 3.78 &4.47 &15.51 \\
\textbf{Turkish} & 82.55 & 62.80 & 87.02 & 66.19 &34.34 & 39.23 &10.39 &7.09 \\
\textbf{Uzbek} & 79.36 & 49.56 & 84.79 & 59.68 &21.84 & 28.83 &4.91 &11.82 \\
\textbf{Yoruba} & 75.75 & 37.13 & 81.34 & 50.72 &19.14 & 25.04 &0.30 &3.21 \\
\midrule
\multicolumn{9}{c}{Out of BERT} \\
\midrule
\textbf{Akan} & 75.87 & 21.96 & 79.33 & 49.02 & 12.82 & 35.2 & 0.52 & 8.42 \\
\textbf{Amharic} & 11.79 & 3.27 & 79.09 & 43.70 & 3.95 & 3.9 & 1.70 & 5.48\\
\textbf{Hausa} & 67.67 & 15.36 & 75.73 & 24.37 & 12.58 & 14.77 &0.19 & 5.64\\
\textbf{Somali} & 74.29 & 18.35 & 84.56 & 53.63 & 15.84 & 21.64 &0.60 & 4.16\\
\textbf{Wolof} & 67.10 & 13.63 & 70.27 & 39.70 & 9.83 & 26.45 &0.09 & 10.63\\
\textbf{Zulu} & 78.89 & 15.82 & 84.50 & 39.65 & 12.3 & 13.72 &0.92 & 11.58\\
\textbf{Uyghur} & 32.64 & 3.59 & 79.94 & 42.98 & 1.45 & 1.52 &1.97 & 2.45\\
\textbf{Tigrinya} & 24.75 & 4.74 & 79.42 & 7.61 & 7.91 & 5.71 &0.01 & 2.20\\
\textbf{Oromo} & 72.00 & 9.34 & 72.78 & 12.28 & 6.84 & 10.11 &0.01 & 2.96\\
\textbf{Kinyarwanda} & 65.85 & 30.18 & 74.46 & 44.40 & 26.55 & 32.3 &0.06 & 0.95\\
\textbf{Sinhala} & 18.12 & 3.43 & 71.63 & 33.97 & 3.39 & 33.97 & 0.10 & 1.02\\
\bottomrule
\end{tabular}
\caption{In the order from left to right, column means: \MBERT with supervision, \MBERT zero-shot cross-lingual, \EBERT with supervision, \EBERT zero-shot cross-lingual. Then we give performance of Hindi and Sinhala \EBERT models when evaluated on all the languages. The last two columns are dataset statistics, with number of million lines in the LORELEI corpus and number of thousand lines in LORELEI NER dataset.}
\label{tbl:performance-comp}
\end{table*}

\begin{table*}[!htbp]
\centering
\begin{tabular}{l l l l l l} 
\toprule
& \multicolumn{2}{c}{\textbf{English}} & \multicolumn{2}{c}{\textbf{Target}}
\\
\cmidrule(l){2-3} \cmidrule(l){4-5}
\textbf{Language}  & \textbf{\EBERT}  & \textbf{ \BBERT} &  \textbf{ \EBERT}  & \textbf{ \BBERT} \\
\midrule
Akan &  79.19 & 77.49 & 49.02  & 48.00\\
Amharic &  78.36 & 78.44 & 43.70 & 38.66\\
Hausa & 74.24 & 80.13 & 24.37 & 26.45\\
Somali & 78.60 & 79.17 & 53.63 & 51.18\\
Wolof &  78.11 & 81.01 & 39.70  & 39.92 \\
Zulu &  79.32 & 81.82 & 39.65 & 44.08\\
Uyghur & 77.76 & 79.65 & 42.98 & 21.94\\
Tigrinya & 76.21 & 80.35 & 7.61 & 6.34\\
Oromo & 76.06 & 78.13 & 12.28 & 8.45\\
Kinyarwanda & 73.05 & 79.37 & 44.4& 46.72 \\
Sinhal &73.70 & 80.04 & 33.97  &16.93  \\
\bottomrule
\end{tabular}
\caption{\textbf{Comparison Between \BBERT vs \EBERT:} We compare the performance of \EBERT with \BBERT on both English and target language. As a reference, performance of \MBERT is 79.37 on English. This shows that neither \BBERT nor \EBERT gets unfair advantage from the English part of the model. }
\label{tbl:bibert-embert-eng-tar}
\end{table*}

\end{document}